\documentclass{bmvc2k}\usepackage{amsmath}
\usepackage{amssymb}
\usepackage{algorithm}
\usepackage{algpseudocode}
\usepackage{wrapfig,lipsum,booktabs}
\usepackage{mathtools} 
\usepackage{booktabs} 
\usepackage{tikz} 
\newcommand\blfootnote[1]{%
  \begingroup
  \renewcommand\thefootnote{}\footnote{#1}%
  \addtocounter{footnote}{-1}%
  \endgroup
}
\title{RISSOLE: Parameter-efficient Diffusion Models via Block-wise Generation and Retrieval-Guidance}

\addauthor{Avideep Mukherjee}{https://www.cse.iitk.ac.in/users/avideep/}{1}
\addauthor{Soumya Banerjee}{https://soubanerjee.github.io/}{1}
\addauthor{Piyush Rai}{https://www.cse.iitk.ac.in/users/piyush/}{1}
\addauthor{Vinay P. Namboodiri}{https://vinaypn.github.io/}{2}
\addinstitution{
 Department of Computer Science and Engineering\\
 Indian Institute of Technology Kanpur\\
 Uttar Pradesh, India
}
\addinstitution{
Department of Computer Science\\
University of Bath\\
Claverton Down\\
Bath BA2 7AY, UK\\
}

\runninghead{Mukherjee, Banerjee, Rai, Namboodiri}{RISSOLE}

\begin{document}

\maketitle

\begin{abstract}
   Diffusion-based models demonstrate impressive generation capabilities. However, they also have a massive number of parameters, resulting in enormous model sizes, thus making them unsuitable for deployment on resource-constraint devices. Block-wise generation can be a promising alternative toward designing compact-sized (parameter-efficient) deep generative models since the model can generate one block at a time, instead of generating the whole image at once.  However, block-wise generation is also considerably challenging because ensuring coherence across the generated blocks can be a non-trivial task. To this end, we design a retrieval-augmented generation (RAG) approach and leverage the corresponding blocks of the images retrieved by the RAG module to condition the training and generation stages of a \emph{block-wise} denoising diffusion model. Our conditioning schemes ensure coherence across the different blocks during training and, consequently, during generation. While we showcase our approach using the latent diffusion model (LDM) as the base model, it can be used with other variants of denoising diffusion models. We validate the solution of the coherence problem through the proposed approach by reporting substantive experiments to demonstrate our approach's effectiveness in compact model size and excellent generation quality.\blfootnote{This research work was partially supported by the Research-I Foundation of IIT Kanpur.}
\end{abstract}
\vspace{-0.5em}
\section{Introduction}\label{sec:intro}
Recent work on deep generative models, particularly diffusion probabilistic models and variants~\cite{ho2020denoising,sohl2015deep,song2020score}, has produced impressive generation quality for data such as images and videos. While recent models can now generate images with very high resolutions, these models also have enormous sizes, which hinders their deployment in resource-constrained settings. Training a state-of-the-art diffusion model from scratch could, for instance, require 23,835 A100 hours (when optimized). \footnote{An example provided in https://www.databricks.com/blog/stable-diffusion-2} Clearly, these models are beyond the computational capabilities of academia or organizations with modest computing resources, which necessitates a parameter-efficient design. 
To address the issue of computationally intensive training and large model sizes, recent work has explored approaches, such as training denoising diffusion models in a lower-dimensional latent space~\cite{rombach2022high}, and using \emph{semi-parametric} retrieval-augmented diffusion models~\cite{blattmann2022semi} where a compact-sized diffusion model is conditioned on an external database. As the training of the diffusion model is done with a fixed latent size, these models still tend to work with large latent representations, resulting in large parameter models that are challenging to train. \emph{Post hoc} attempts at optimization such as adversarial diffusion distillation \cite{sauer2023adversarial} are also not applicable as one still needs access to a fully trained large parameter model and further resources to train a distilled model.

Training deep generative models such that they can generate images in a \emph{block-wise} fashion (generating one block at a time), instead of the whole image at once, can potentially be an alternative to address the aforementioned issues. However, designing such block-wise generative models is quite challenging, as ensuring that the generated blocks have spatial and semantic coherence is difficult. Consequently, the generated blocks may not blend well together, resulting in noticeable discrepancies or artifacts where the blocks meet.

In this work, we show that the retrieval-augmented generation (RAG) approach~\cite{blattmann2022semi} can effectively address the consistency problem in block-wise generation. In particular, we present a block-wise denoising diffusion model where each image block (in the pixel space, or the latent space if using a latent diffusion model~\cite{rombach2022high}), both during training as well as generation, is conditioned on the corresponding block of a reference set of images retrieved from an external database. We show that enforcing correspondence in the block-wise conditioning suffices to obtain coherence in the \textit{independent} block-wise generation.


\vspace{-1em}
\section{Related Work}
There is extensive work on deep generative models for data, such as images, videos, text, and multi-modal data. Here, we mainly focus on generative models based on diffusion probabilistic models~\cite{sohl2015deep} and their variants \cite{kong2021fast,san2021noise,song2020denoising,dhariwal2021diffusion,ho2020denoising,ronneberger2015u,song2020score}, particularly those focusing on compact model sizes, as they are most relevant to our work. There has been an extensive exploration of variants of diffusion-based models, such as score-based models and their application on various data sources \cite{diff_survey}. 

Our work mainly focuses on designing diffusion models that are parameter-efficient \emph{by design} and can consequently be trained efficiently. There are three main strategies: optimizing the representation (using probability flow ODEs), training using a latent representation, and patch-based training. Models such as consistency models \cite{song2023consistency} obtain an approximation for the stochastic differential equation based on the diffusion process. However, the main contribution of this approach is to reduce the number of function evaluations required at inference time. This approach still requires substantial training time. Methods such as the latent diffusion model (LDM) \cite{rombach2022high} model the diffusion process in the latent space instead of the original image space. One could theoretically use a highly compact latent space to reduce the computational cost. However, as the efficacy of the encoder-decoder transformation is limited, typically, one uses a sizeable latent space representation while training these models, and these LDMs require thousands of hours of large GPU-based training. 

Patch-based training of diffusion models is an alternative. For instance, \cite{wang2023patch} shows that training on patches could reduce the training time. However, they do not solve the coherence issue. Their strategy is to train with various patches of various sizes and to train using a few \emph{full-sized} images during training, thus not yielding any model-sized improvements. Another recent patch-based strategy~\cite{ding2024patched} trains with a patch-based model. However, they provide the context from the feature collage of patches to appropriately condition the image generation. We do evaluate a variant of this approach where we condition on adjacent blocks. We observed that the representation does not train well with our proposed strategy.

Our work is based on more explicitly and directly incorporating coherence by conditioning the individual blocks on the corresponding blocks of "relevant" images retrieved from an external database using the idea of retrieval-based augmentation. Our work in incorporating coherence also has relevance to the broader area of coherence in generative models \cite{xu-etal-2018-skeleton, chu2020learning}. Our proposed approach is to develop block-based retrieval-guided diffusion models. This is complementary to other approaches of probability flow ODE \cite{song2023consistency} or latent diffusion \cite{rombach2022high} and can be combined with these different approaches.  
\vspace{-1em}
\section{Background}
\label{sec:background}
\textbf{Diffusion Models:} Denoising diffusion probabilistic models (DDPM)~\cite{ho2020denoising,sohl2015deep,song2020score} are deep generative models that consist of a forward process and a reverse process. The forward process $q(x_{1:T} \vert x_0) = \prod_{t=1}^Tq(x_t \vert x_{t-1})$ takes a clean input $x_0$ and gradually corrupts it by producing a sequence $x_1,x_2,\ldots,x_T$ defined by forward diffusion of the form $q(x_t|x_{t-1})$, usually modeled by a Gaussian $q(x_t \vert x_{t-1}) = \mathcal{N}(\sqrt{1-\beta_t}x_t, \beta_t\mathbf{I})$. Note that $q(x_t \vert x_0) = \mathcal{N}(\sqrt{\bar{\alpha_t}}x_0, (1-\bar{\alpha_t})\mathbf{I})$, where $\alpha_t = 1 - \beta_t$ and $\bar{\alpha_t} = \prod_{t}\alpha_t$. For a sufficiently large value of $T $ and a suitably chosen variance schedule $\{\beta_t\}_{t=1}^T$, the distribution $q(x_T  \vert x_0)$ approximates an isotropic Gaussian. A \emph{learnable} reverse process, called reverse denoising, denoted as $p(x_{0:T}) = p(x_T)\prod_{t=1}^Tp_\theta(x_{t-1} \vert x_t)$, tries to reconstruct the clean data $x_0$ from pure noise $x_T$ with $p(x_T) = \mathcal{N}(0, \mathbf{I})$ and $ p_\theta(x_{t-1} \vert x_t) = \mathcal{N}(\mu_\theta(x_t,t), \Sigma_\theta(x_t, t))$,
where the parameters $\mu_\theta(x_t,t), \Sigma_\theta(x_t, t)$ of the denoising network are defined using a deep neural network. Once learned, the reverse process can synthesize new inputs from pure noise $x_T$. It can also be shown that estimating the reverse process parameters is equivalent to predicting the noise in the distribution $q(x_t  \vert x_0)$.

\noindent \textbf{Latent Diffusion Models:} While the forward and reverse diffusion processes of the standard DDPM operate in the image space, the latent diffusion model (LDM)~\cite{rombach2022high} takes a different approach where both of these processes operate in a low-dimensional latent space. The LDM consists of an encoder $\mathcal{E}$ which compresses the input image $x$ into a lower-dimensional latent representation $z$, a decoder $\mathcal{D}$ that reconstructs $x$ from $z$, and the forward and reverse diffusion processes operate on the latent $z$ representation. In addition to yielding a semantically meaningful latent space, LDM has immediate benefits regarding reduced model size, training cost, and sample generation speed.

\noindent \textbf{Retrieval-Augmented Generation:} An orthogonal approach to designing compact-sized deep generative models is retrieval-augmented generation (RAG)~\cite{zhao2024retrieval} which augments a compact generative model with an external database \(\mathcal{D}\). RAG uses similarity-based retrieval during training/generation to condition the generative model based on additional data retrieved from this external database. Blatmann et al. \cite{blattmann2022semi} leveraged the RAG idea for diffusion models by employing a non-parametric retrieval strategy \(\xi_k: \mathcal{D} \mapsto \mathcal{M}_\mathcal{D}^{(k)}\) such that \(\mathcal{M}_\mathcal{D}^{(k)} \subseteq \mathcal{D}\) and \( \vert \mathcal{M}_\mathcal{D}^{(k)} \vert =k\). During the training phase, for each sample \(x \sim p(x)\), their proposed retrieval-augmented diffusion model (RDM) uses \(\xi_k\) to acquire a set of \(k\) nearest neighbors \(\mathcal{M}_\mathcal{D}^{(k)}(x)\) from the database \(\mathcal{D}\). Subsequently, the generative model is conditioned on the representations of these retrieved instances, eliminating the necessity to synthesize an image region entirely from its parameter manifold. While RAG is primarily motivated by the need for compact models, our work shows how RAG can also be leveraged to solve the challenging problem of ensuring coherence across blocks when doing block-wise generation.
\vspace{-1em}
\section{Denoising Diffusion Models with Block-wise Generation}
\begin{figure}[!t]
    \centering
    \includegraphics[width=\linewidth, keepaspectratio]{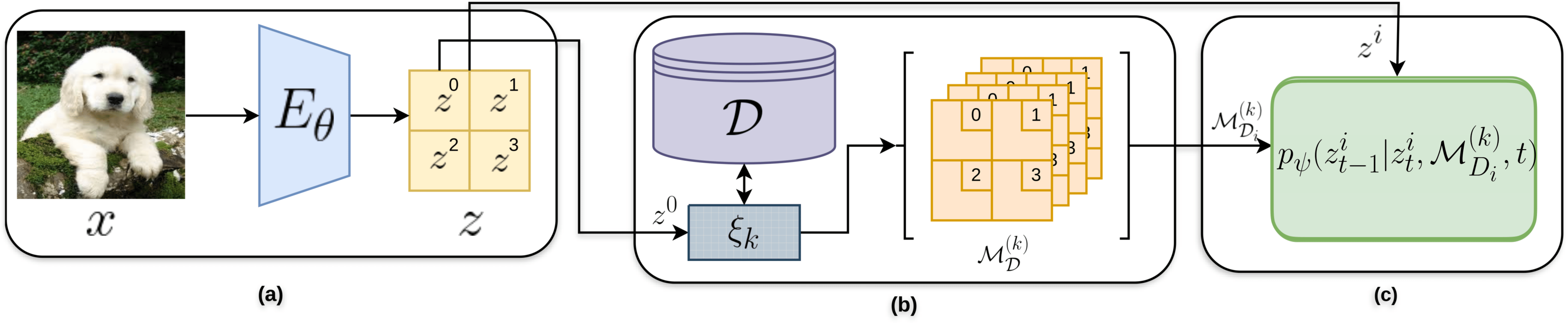}
    \includegraphics[width=\linewidth, keepaspectratio]{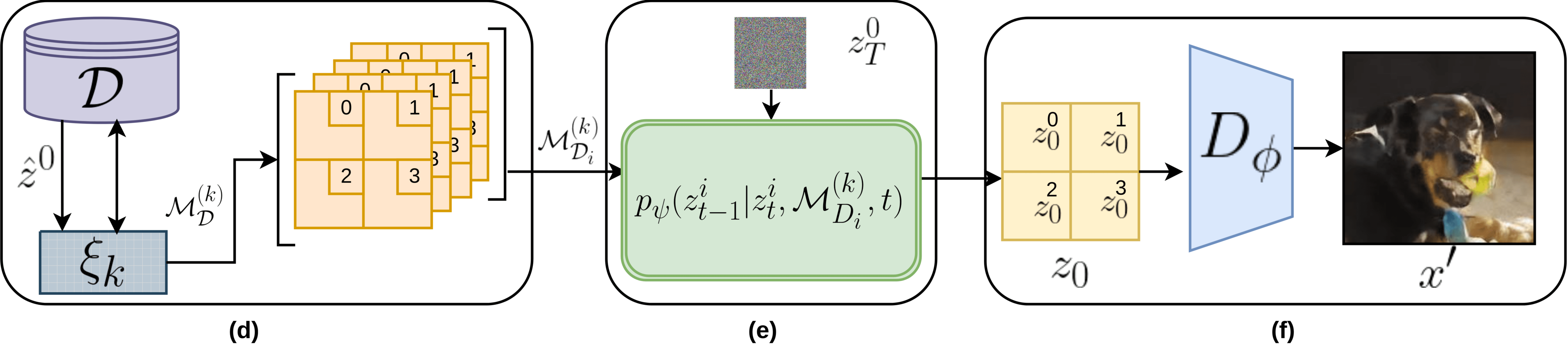}
    \vspace{-1em}
    \caption{\small{Training and Sampling of RISSOLE: (a) During training, each image \( x \) goes through a VQ-GAN encoder \( E_\theta \), resulting in the generation of a latent representation \( z \). (b) This latent representation \( z \) is then utilized by the retriever \( \xi_k(.) \) to fetch its \( k \) nearest neighbors \( \mathcal{M}_{\mathcal{D}}^{(k)} \) from a dataset \( \mathcal{D} \). (c) Each group of blocks \( \mathcal{M}_{\mathcal{D}_i}^{(k)} \in \mathcal{M}_{\mathcal{D}}^{(k)} \) is used as a conditioning signal for the corresponding \( z^i \), aiding in the training procedure. (d) During the sampling (generation) phase, a pseudo-query \( \hat{z}^0 \) obtained from the dataset \( \mathcal{D} \) is employed by \( \xi_k(.) \) to retrieve \(\mathcal{M}_{\mathcal{D}}^{(k)} \). (e) From the retrieved set \( \mathcal{M}_{\mathcal{D}}^{(k)} \), each subset \( \mathcal{M}_{\mathcal{D}_i}^{(k)} \) is used as a conditioning signal for the random noise \( z^i_t \) at steps $t=T,T-1,\ldots,3,2,1$ to generate the final denoised block \( z_0^i \) of \( \hat{z}^0 \). (f) All denoised representations \( z_0^i \) are reshaped to construct \( z_0 \), which is then passed through the decoder \( D_\phi \) to yield the reconstructed sample \( x' \).}}
    \label{fig:model_train}
    \vspace{-1em}
\end{figure}

Our block-wise generation approach applies to DDPMs operating in the image space and those operating in a latent space. This work will illustrate our approach using the latent diffusion model (LDM)~\cite{rombach2022high}, an example of the latter class of methods. Our approach (Fig.~\ref{fig:model_train}) assumes that the input to each forward/reverse diffusion step is partitioned into $b$ disjoint blocks. Omitting the time-step subscript $t$ and denoting the latent representation at any time-step as $z$, we assume it to be partitioned into $b$ equal-sized disjoint blocks as $z = \bigcup_{i=0}^{b-1} z^i$. Operating on smaller-sized blocks $z^i$ naturally results in a model with fewer parameters, which is our primary goal. However, the challenge here is to ensure coherence among the blocks (e.g., avoiding artifacts at the block boundaries, semantic inconsistency between adjacent blocks, etc.). To achieve coherence, each block $z^i$ is also conditioned on the \textit{corresponding} block of a set of $k$ nearest neighbors retrieved from the external database; Sec.~\ref{sec:rissole_training} provides the details. We refer to our approach as RISSOLE (Paramete\textbf{R}-effic\textbf{I}ent Diffu\textbf{S}ion Models via Block-wi\textbf{S}e Generati\textbf{O}n and Retrieva\textbf{L}-Guidanc\textbf{E}).
\vspace{-1em}
\subsection{RISSOLE Training}
The training procedure of RISSOLE consists of the following steps: (1) VQ-GAN encoder-decoder training because our model operates in a latent space,(2) Building the database from which retrieval is performed, and (3) DDPM training in the latent space using a block-wise generation approach and leveraging RAG to ensure coherence across blocks. Next, we describe each of these steps in more detail.

\vspace{-0.5em}
\subsubsection{VQ-GAN Training} 
\label{sec:VQ-GAN}
As the first stage of RISSOLE training, we train a VQ-GAN on the original training images to obtain their latent representations using a trainable encoder-decoder setup. For an image \(x \in \mathcal{X} \subset \mathbb{R}^{h \times w \times c}\), we introduce an encoder \(E_\theta\), where \(E_\theta\) takes \(x\) as input and produces a hidden representation \(z \in \mathbb{R}^{h' \times w' \times c'}\). Here, \(h, w, c\) represents the dimensions of the input image, \(h', w', c'\) represents the dimensions of the hidden representation, and \(\theta\) represents the parameters of the encoder \(E\) s.t. $z = E_\theta(x)$. Similar to LDM)~\cite{rombach2022high}, and to be amenable to our subsequent block division approach, we maintain $z$ in a two-dimensional layout, departing from the arbitrary one-dimensional arrangement used in the works of \cite{esser2021taming} and \cite{ramesh2021zero}. This reduces the image space by a factor of \(f = w/w'\) or \(h/h'\). The factor \(f\) determines the size of each block into which the hidden representations are divided. A decoder \(D_\phi\), parameterized by \(\phi\), reconstructs the original image as $\tilde{x} = D_\phi(E_\theta(x)) \in \mathbb{R}^{h \times w \times c}$.
\vspace{-0.5em}
\subsubsection{Building the Retrieval Database} 

For constructing the retrieval database, denoted as $\mathcal{D}$, the latent representations $z$  of the training images are partitioned into $b$ non-overlapping, equal-sized blocks. We denote $\mathcal{D}$ as $\bigcup_{i=0}^{b-1} \mathcal{D}_i$, where $\mathcal{D}_i$ comprises a stack of the $i^{th}$ blocks of $z = E(x)$ for all $x \in \mathcal{X}$. 
\vspace{-0.5em}
\subsubsection{Block-wise DDPM Training}
\label{sec:rissole_training}
Recall that the forward and reverse diffusion of our proposed DDPM formulation operate on the latent representation $z = \bigcup_{i=0}^{b-1} z^i$. Before iterating through each block, $z^i$, we use the first block\footnote{Another design choice can be to use the entire $z$ to do the retrieval since we have access to the full image and its latent representation. We show some results with this choice in the Appendix.} $z^0$ to retrieve the $k$ nearest neighbors of $z^0$ from $\mathcal{D}$, denoted by $\mathcal{M}_{\mathcal{D}}^{(k)} = \xi_k(z^0, D)$, where $\xi(.)$ is the retrieval function. Note that once $\mathcal{M}_{\mathcal{D}}^{(k)}$ is obtained,  we do not need to perform the retrieval operation again to obtain $\mathcal{M}_{\mathcal{D}_i}^{(k)}$ for subsequent blocks $z^i$  for $i \in {0, 1, 2, \cdots, b-1}$, since $\mathcal{M}_{\mathcal{D}}^{(k)} = \bigcup_{i=0}^{b-1}\mathcal{M}_{\mathcal{D}_i}^{(k)}$. Subsequently, each block $z^i$ is then conditioned on $\mathcal{M}_{\mathcal{D}_i}^{(k)}$ to facilitate the training process through the DDPM U-Net, parameterized by $\psi$. This approach, characterized by its granularity and reliance on semi-parametric generative modeling~\cite{blattmann2022semi}, can be expressed as
\vspace{-0.5em}
\begin{equation}
    p_{\psi, \mathcal{D}, \xi_k}(z) = \prod_{i=1}^{b-1}p_\psi(z^i  \vert  \{y  \vert  y \in \mathcal{M}_{\mathcal{D}_i}^{(k)}\})
    \vspace{-0.5em}
\label{eqn:model}
\end{equation}
The neighboring information denoted as $\mathcal{M}_{\mathcal{D}_i}^{(k)}$, as retrieved from the database $\mathcal{D}_i$, is typically combined with the input block $z^i$ by simply putting them together along the channel dimension. This combined information is input into a U-Net architecture~\cite{ronneberger2015u} during the reverse denoising stage within a latent diffusion model. However, we take a different approach where the input and the retrieved neighbors are processed separately through different convolutional layers, resulting in distinct standard embeddings. These embeddings are added to the original input before feeding into the U-Net structure.  To maintain consistency and prevent unwanted shifts during training, we apply Layer Normalization \cite{ba2016layer} to the output. This modification allows us to forgo the costly Cross Attention Mechanism \cite{rombach2022high} for conditioning the input, resulting in fewer parameters and faster training and sampling times.

We train RISSOLE using the standard re-weighted likelihood objective, as described in \cite{ho2020denoising}, resulting in the objective similar to the one used in \cite{ho2020denoising,sohl2015deep}:
\vspace{-0.5em}
\begin{equation}
    \min_{\psi}\mathcal{L} = \sum_{i=0}^{b-1}\mathbb{E}_{p(x),z \sim E_\theta(x), \epsilon \sim \mathcal{N}(0, 1), t}[|| \epsilon - \epsilon_\psi(z^i_t, t, \{y  \vert  y \in \mathcal{M}_{\mathcal{D}_i}^{(k)}\}) ||]
    \label{eq:rissole_objective}
\end{equation}

In Eq.~\ref{eq:rissole_objective}, $\epsilon_\psi$ denotes the UNet-based denoising autoencoder detailed in \cite{ronneberger2015u} and applied in \cite{rombach2022high,dhariwal2021diffusion}. In addition to serving as the objective for a typical DDPM model, Eq.~\ref{eq:rissole_objective} also includes the collection of retrieved neighboring data points as a prerequisite for the noise predictor U-Net: $\epsilon_\psi$. Furthermore, the loss is summed across all blocks where it is individually computed. The notation $t \sim \text{Uniform}\{1, \ldots, T\}$ indicates the time step, as elaborated in \cite{sohl2015deep,rombach2022high}. A pseudocode outlining the training procedure can be found in the Appendix.

\subsection{RISSOLE Sampling}
Like RDM~\cite{blattmann2022semi}, RISSOLE begins by picking a random \emph{pseudo-query} from the database for generating new samples. However, unlike RDM, which uses a whole image as the query, we use its first block $\hat{z}^0$ from the first block of our database $\mathcal{D}_0$. We then use the query $\hat{z}^0$ to retrieve $k$ nearest neighbors from the database (see Fig.~\ref{fig:model_train} (d)).

Thereafter, we generate a random noise sample from a standard Gaussian. This sample is divided into $b$ fixed-size blocks, each covering a part of the latent space. Next, we perform block-wise sampling, where each block $z^i$ is sampled iteratively conditioned on its $k$ nearest neighbors $\mathcal{M}_{\mathcal{D}_i}^{(k)}$ from the database, following a process similar to the one used during training. Once all blocks have been updated, we reconstruct the complete latent representation by combining the representations from each block. Finally, we decode this combined latent representation using the VQ-GAN Decoder $D_\phi$ with parameters $\phi$ to produce a sample $x$ in the original data space. This enables structured sampling from the latent space, capturing relationships within blocks while maintaining coherence across generated samples. 

A distinguishing aspect of RISSOLE is that both training and sampling can be parallelized across the $b$ blocks (as the conditioning structure we use does not depend on the blocks on each other), thereby yielding further speed-ups during training and sampling. A pseudocode for sampling from RISSOLE is presented in the Appendix.

\vspace{-1em}
\section{Experiments}
\subsection{Dataset and Baselines}
We evaluate RISSOLE on two benchmark datasets: CelebA $64 \times 64$ \cite{liu2015faceattributes} and ImageNet100 \cite{deng2009imagenet}. CelebA64 contains 202,599 images showing human faces with $64 \times 64 \times 3$ resolution, while ImageNet $100$ has 127,878 natural images from 100 categories, each with a resolution of $224 \times 224 \times 3$. We used 80\% of the data for training and the remaining 20\% for validation, and the same training/validation split was used for both datasets. We compare RISSOLE with RDM~\cite{blattmann2022semi}, a recent SOTA method for compact-sized diffusion models. RDM uses an external database to keep the size of the diffusion model module small. In addition to RDM, we also compare RISSOLE with two other baselines - a variant of RISSOLE that does not use retrieval guidance and another variant that does not use retrieval guidance but uses the different blocks of the image to condition the current block (this is an instance of the recent work on patch-based diffusion models~\cite{ding2024patched}). Please note that these are the most relevant baselines for RISSOLE. Although fast to train, recent methods like patch diffusion~\cite{ding2023patched} are not directly comparable with RISSOLE since such approaches still require a small set of full-sized images during training.
Nevertheless, we compare RISSOLE with Patch Diffusion when it does not use full-sized images during training to ensure fairness in comparison. Thus, their model sizes are not small, unlike RISSOLE, and are thus not included in the comparison here. It is also important to highlight that the roles played by the external database differ in RDM and RISSOLE. In RDM, it helps keep the diffusion model size small, whereas in RISSOLE, the primary role of the external database is to achieve coherence across blocks. In contrast, RISSOLE's block-wise training/generation ability contributes to its parameter efficiency.

\subsection{Experimental Setup}
In training the VQ-GAN encoder-decoder, we have maintained $f=8$ (refer to Section \ref{sec:VQ-GAN}) across both datasets. Specifically, for the CelebA dataset, the VQ-GAN's latent dimension is configured to 10. Conversely, due to ImageNet's higher resolution, the latent dimension is adjusted to 32. Consequently, the latent resolution for an image from CelebA will be $10 \times 16 \times 16$, while for ImageNet, it will be $32 \times 28 \times 28$. To retrieve the nearest neighbors from $\mathcal{D}$, we use ScanNN \cite{guo2020accelerating} search algorithm in the latent representation generated by a VQ-GAN \cite{esser2021taming}.

\begin{wrapfigure}{r}{0.55\textwidth}
    \includegraphics[width=\linewidth, keepaspectratio]{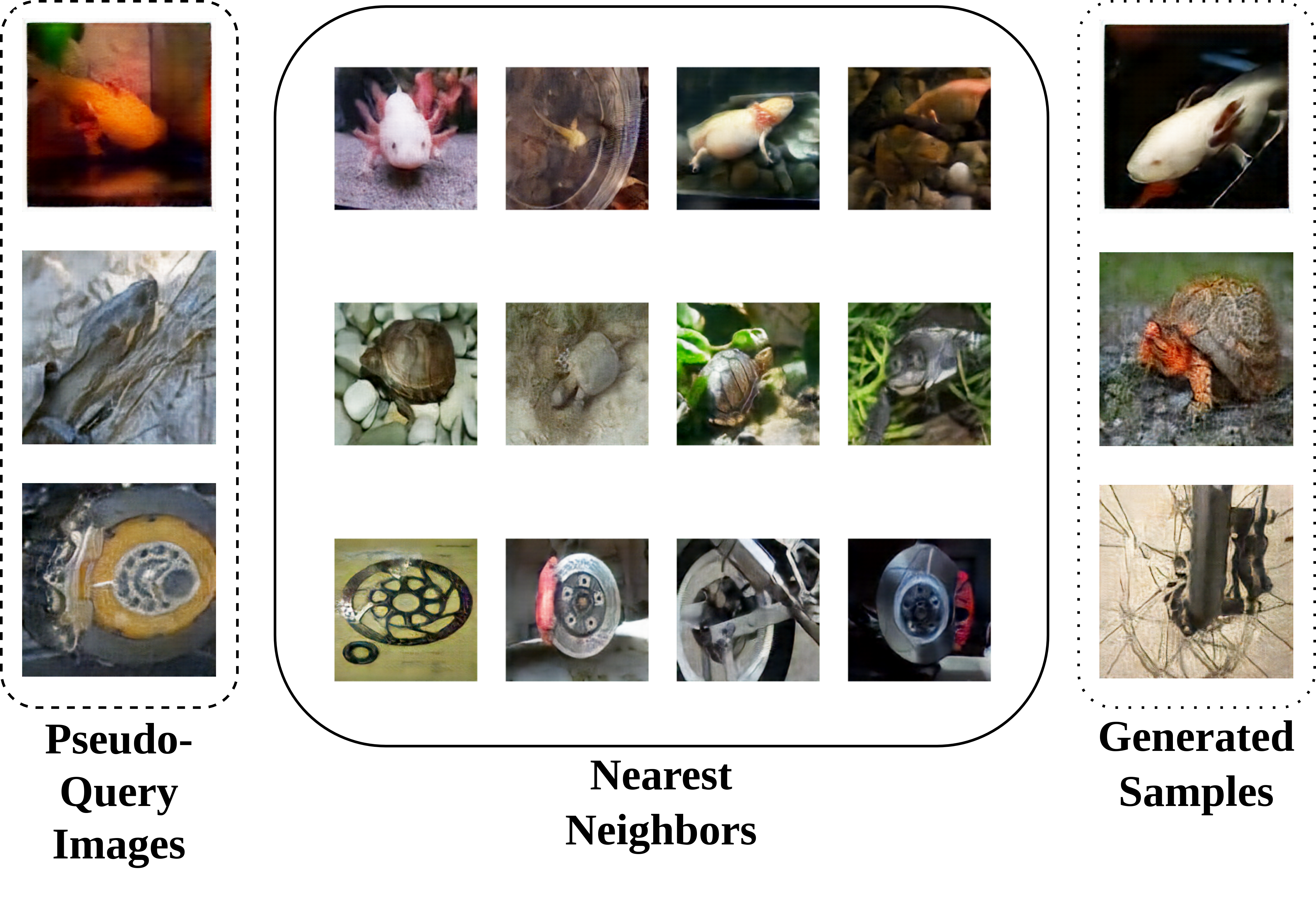}
    \vspace{-2em}
    \caption{\small{Each of the three rows above show a pseudo-query image $\hat{x}$ (used in generation time) from $\mathcal{D}$, its retrieved neighbors, and the generated sample when these neighbors are conditioned on. Note that the actual training and sampling occur in the latent space. These images, decoded from the latent representations, are for better understanding and visualization.}}
    \vspace{-0.8em}
    \label{fig:fid_k}
\end{wrapfigure}

In alignment with the requirements of the chosen retrieval mechanism, each latent representation is flattened. For instance, considering the ImageNet100 dataset $\mathcal{X} \subset \mathbb{R}^{224 \times 224 \times 3}$ with $ \vert \mathcal{X} \vert  = n$, and $z \in  \mathbb{R}^{28 \times 28 \times 32}$ the dimensionality of $D$ is denoted as $\mathbb{R}^{b \times n \times 6272}$, with each $\mathcal{D}_i \in \mathbb{R}^{n \times 6272}$ where $6272$ emerges from the flattening process applied to $z$. Our models are trained for 200 epochs on an NVIDIA 1080Ti (CelebA64) and an NVIDIA A30 (ImageNet100). The number of nearest neighbors $k$ can influence the performance of retrieval-based approaches. In our experiments, $k=20$ was found to work best for RDM~\cite{blattmann2022semi} and $k=10$ worked best for RISSOLE. In addition, we also conduct an ablation experiment in Sec.~\ref{sec:pos} to assess RISSOLE's performance when conditioned with positional information.

\subsection{Unconditional Image Synthesis}

Upon generating samples from our diffusion model, we observed a notable improvement in visual quality compared to the RDM baseline when constrained to roughly the same model size as RISSOLE. Despite its compact model size, the images generated by RISSOLE exhibit finer details, sharper features, and more realistic textures. This qualitative assessment is further supported by quantitative metrics, including the Fréchet Inception Distance (FID).

\begin{figure}
   \centering
     \includegraphics[width=0.7\linewidth]{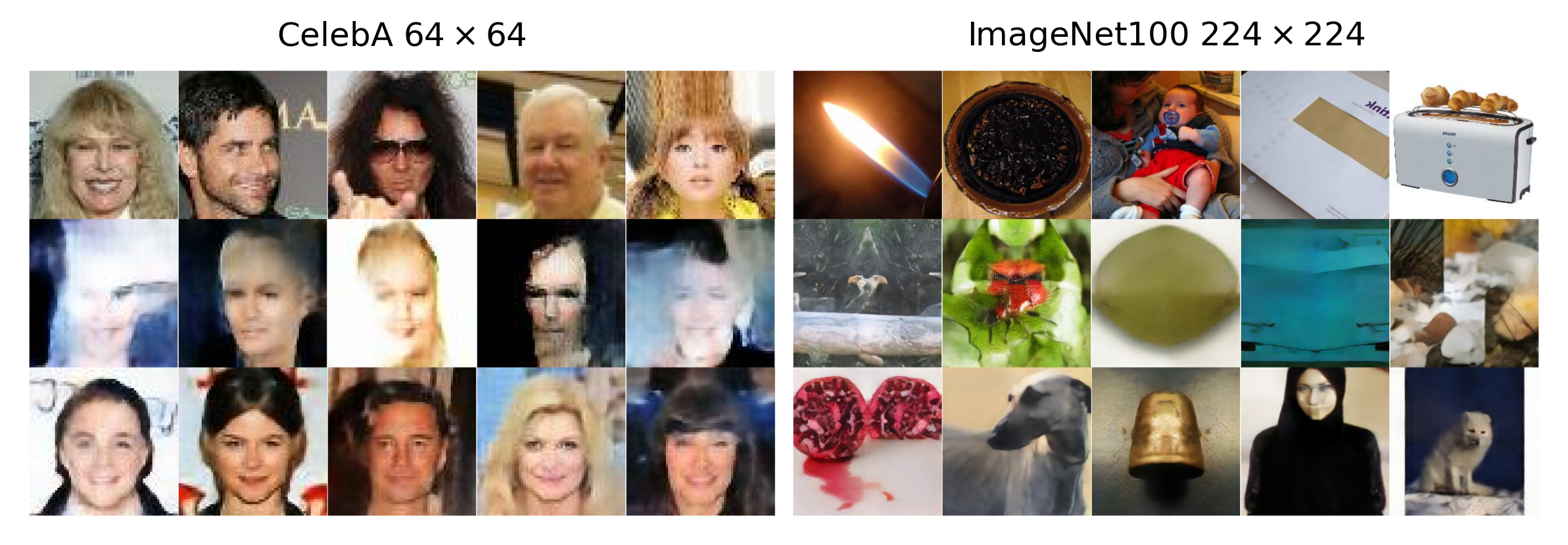}
     \vspace{-1em}
     \caption{{Original Images (top row), and samples generated by the RDM baseline (middle row) and by RISSOLE (bottom row), trained on CelebA and ImageNet 100 datasets.}}
     \label{fig:samples_comparison}
     \vspace{-1em}
\end{figure}

Table \ref{tab:comparison} presents the FID scores obtained by RISSOLE and the other baseline models on both the CelebA and ImageNet datasets. RISSOLE achieves significantly lower FID scores, indicating that its generated samples are more likely to be from the original dataset distribution. These results demonstrate that RISSOLE captures the intricate patterns and structures in the data and generates more realistic and coherent images than the other baselines with similar model sizes. To further illustrate the qualitative differences, Fig.~\ref{fig:samples_comparison} compares samples generated by RISSOLE, original images from both datasets and samples from the RDM baseline. As depicted in Fig.~\ref{fig:samples_comparison}, RISSOLE produces noticeably better samples, closely resembling the characteristics of the original images. For fairness of comparisons, we aimed to ensure that the model sizes for RDM and RISSOLE are kept similar, as reported in Table~\ref{tab:comparison}. To ensure roughly equal model sizes for RDM and RISSOLE, the U-net architecture used by RDM was suitably modified. However, note that because RDM still utilizes the entire image or latent space for training or sampling, it still requires a somewhat higher number of parameters under this constrained setting. Table~\ref{tab:comparison} shows that RISSOLE outperforms RDM on both datasets and Patch Diffusion on CelebA $64 \times 64$ \footnote{FID values for Patch Diffusion are from the original paper. No FID results for ImageNet100 were reported.}. When constrained to small model size, RDM struggles to model the full-sized images with high fidelity and faces difficulty producing coherent and interpretable sample outputs, as shown in the middle row of Fig.~\ref{fig:samples_comparison}. In contrast, RISSOLE operates only on small-sized blocks and can, therefore, model these blocks more effectively despite its small model size.
\begin{table}[ht]
    \centering
    \caption{Comparison of RISSOLE with RDM\cite{blattmann2022semi}, Patch Diffusion\cite{wang2023patch} and other variants of RISSOLE for unconditional image generation on ImageNet100 \cite{deng2009imagenet}.}

    \footnotetext{FID values for Patch Diffusion are from the original paper. No FID results for ImageNet100 were reported.}

    \label{tab:comparison}
    \vspace{0.2em}
    \begin{tabular}{c|c|c}
    \toprule
    \textbf{Model} & \textbf{CelebA} & \multicolumn{1}{l}{\textbf{ImageNet100}} \\
    \midrule
    RDM & 32.36 & 60.61 \\
    Patch Diffusion & 14.51 & - \\
    RISSOLE - RAG & 23.45 & 29.59 \\
    RISSOLE + P & 17.01 & 20.28 \\
    RISSOLE + Pos & 11.61 & 14.59 \\
    RISSOLE & \textbf{9.82} & \textbf{12.93}\\
    \bottomrule
    \end{tabular}

    \vspace{-1em}
\end{table}

\vspace{-1em}
\subsection{Ablation}
We also conduct ablation studies to gain insights into the behavior and performance of RISSOLE under different configurations and settings. Specifically, we investigate the impact of varying the number of nearest neighbors retrieved from the database on the model's performance, measured in terms of FID scores. We also vary the number of blocks into which the latent space should be divided to observe how the sampling quality changes. Furthermore, we discuss the impact of the retrieval augmented generation to empirically demonstrate how it helps in the coherence of the generated images.



\subsubsection{Impact of Positional Information}\label{sec:pos}
In Section \ref{sec:rissole_training}, we described how RISSOLE learns a specific segment of the latent space, \( z^i \), by conditioning it with \( \mathcal{M}_{\mathcal{D}_i}^{(k)} \), while also integrating positional information, represented by \( i \). This section delves into the significance of positional information within RISSOLE, suggesting that such information might already be inherent in \( p_\psi(z^i|\mathcal{M}_{\mathcal{D}_i}^{(k)}) \). Hence, the model can represented by a modified version of Equation \ref{eqn:model}:

\begin{equation}
    p_{\psi, \mathcal{D}, \xi_k}(z) = \prod_{i=1}^{b-1}p_\psi(z^i  \vert  \{y  \vert  y \in \mathcal{M}_{\mathcal{D}_i}^{(k)}\}, i)
    \vspace{-0.5em}
\label{eqn:model_with_pos}
\end{equation}

\begin{figure}
    \centering
    \includegraphics[width=0.5\textwidth, keepaspectratio]{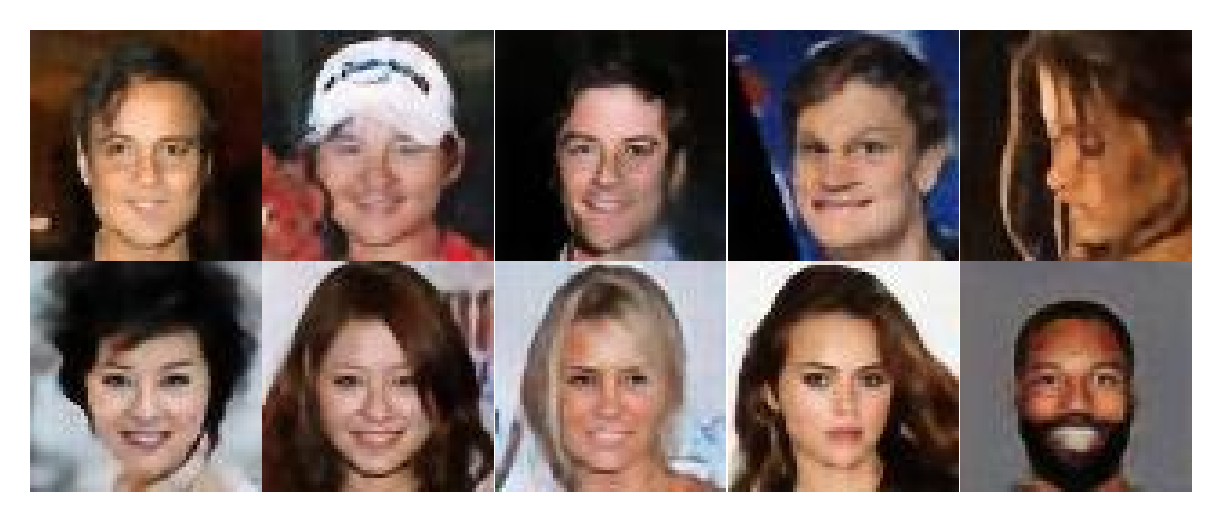}
    \caption{Qualitative Samples from RISSOLE models where the input is conditioned with (top) and without (bottom) the positional information.}
    \label{fig:posless}
\end{figure}

The experiments are conducted using the CelebA dataset with a block count of \( b=4 \). Figure \ref{fig:posless} demonstrates two sets of samples generated by RISSOLE; the upper row displays samples with included positional information, while the lower row exhibits samples without positional information. Observation of the figure visually confirms that samples generated from RISSOLE without positional conditioning exhibit sharper and more realistic characteristics compared to those with positional encoding. This phenomenon arises from the additional complexity introduced by positional encoding, making it more challenging for the model to estimate the underlying data distribution accurately. While the CelebA samples generated with positional conditioning yield an FID score of 11.93, as reported in Table \ref{tab:comparison} RISSOLE without positional encoding surpasses this with an FID of \textbf{9.82}. This finding substantiates our conjecture that retrieval-augmented generation and block-wise conditioning inherently preserve positional information, consequently enabling RISSOLE to be trained without explicit positional conditioning, thereby enhancing the model's time efficiency.


\begin{wrapfigure}{r}{0.35\textwidth}
\vspace{-3em}
    \includegraphics[width=\linewidth, keepaspectratio]{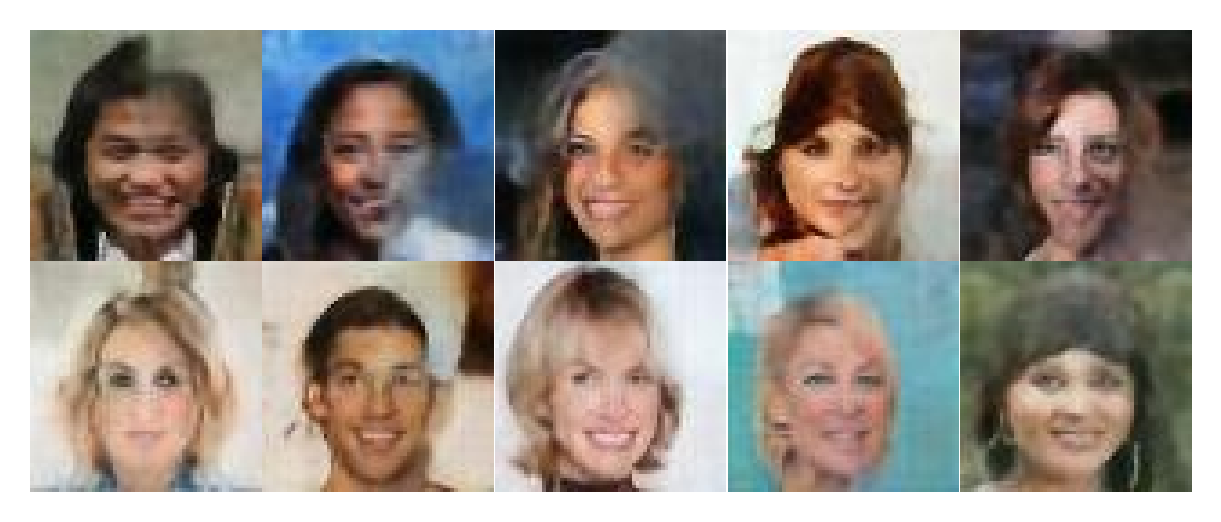}
    \vspace{-1em}
    \caption{\small{Samples from RISSOLE with (top) and without(bottom) using the previous block as a condition.}}
    \vspace{-1em}
    \label{fig:prev}
\end{wrapfigure}
\subsubsection{Using Previous Block as Additional Context}
We also explore conditioning on the previously generated block \emph{in addition to} the nearest neighbors from the external database and compare the model's performance with and without this extra conditioning to assess its impact on image synthesis quality.
Figure \ref{fig:prev} indicates that \emph{also} employing the previous block in conditioning results in a degradation in image quality compared to when it is not used. Specifically, samples generated without utilizing the previous block exhibit finer details, sharper features, and higher fidelity, indicating superior visual quality. FID scores of RISSOLE with the previous block serving as a context (RISSOLE-P) are presented in Table \ref{tab:comparison}. Incorporating the previous block as a condition may impose limitations restricting the model's capacity to comprehend the complete diversity and intricacy of the underlying data distribution. Conversely, training the model without any context from the preceding or other blocks enables the parallelization of the training process, resulting in enhanced speed.
\vspace{-1em}
\subsubsection{Impact of RAG}
\begin{wrapfigure}{R}{0.35\textwidth}
    \includegraphics[width=\linewidth, keepaspectratio]{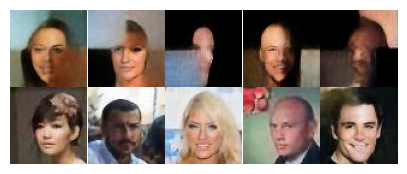}
    \vspace{-2em}
    \caption{\small{RISSOLE samples with (bottom) and without (top) using RAG}}
    \vspace{-1em}
    \label{fig:ragless}
\end{wrapfigure}
We examine the role of the retrieval augmented generation (RAG) mechanism in our diffusion model by comparing the generated samples with and without RAG in Figure \ref{fig:ragless}. When the RAG mechanism is omitted, we observe a degradation in image quality, resulting in incoherent and less visually appealing samples. Table \ref{tab:comparison} shows the FID scores of a RISSOLE model trained without RAG conditioning, which quantitatively confirms the same. 

Incorporating the RAG mechanism is crucial in guiding the block-wise diffusion process and ensuring the coherence and realism of generated images. The retrieved neighbors offer a comprehensive view of the image, aiding the model in generating coherent images block-by-block without needing explicit information about previous blocks. Conditioning the noise on retrieved nearest neighbors through RAG ensures consistency and structural integrity, enhancing the overall fidelity of the synthesized images.
\section{Conclusion}

In this paper, we presented a block-based retrieval-augmented generation (RAG) as an effective solution for the growing model size of denoising diffusion models. The inclusion of RAG addresses any consistency issues in block-wise image generation. We demonstrate improved spatial and semantic coherence in the generated images by leveraging RAG within a block-wise denoising diffusion model. Our approach, which conditions each image block on a corresponding block from a reference set retrieved from an external database, effectively mitigates the challenges associated with block-wise generation. Our experiments and analysis show that RAG can enhance the quality of block-wise generated images while reducing computational complexity and model size. This approach offers a promising alternative for resource-constrained settings where deploying large generative models may not be feasible. Further research can explore extensions and optimizations of RAG-based block-wise generation methods. Additionally, investigating the applicability of our approach to other domains beyond image generation could uncover new opportunities for leveraging retrieval-augmented techniques for enhanced generative modeling.

\bibliography{egbib}
\end{document}


\maketitle
\vspace{-3em}
\section{Algorithms for Training and Sampling of RISSOLE}
\vspace{-1em}
\begin{algorithm}
    \caption{RISSOLE Training}
    \label{algo:train}
    \begin{algorithmic}[1]
        \Repeat
            \For{$x \sim q(x)$}
                \State $z = E_\theta(x)$ 
                \State Divide $z$ into $b$ fixed-size blocks $\ni \bigcup_{i=0}^{b-1} z^i = z$
                \State $z^0 \sim \mathcal{N}(0,\mathbf{I})$
                \State $\mathcal{M}_{\mathcal{D}}^{(k)} = \xi_k(z^0, \mathcal{D})$ where $\mathcal{M}_{\mathcal{D}}^{(k)} = \bigcup_{i=0}^{b-1}\mathcal{M}_{\mathcal{D}_i}^{(k)}$
                \For{each $z^i$}
                    \State $\epsilon = q(z_t^i \vert z^i)$ for any $t \sim \text{Uniform}(1, \cdots, T)$
                    \State $\epsilon'_\psi = p_\psi(z_{t-1}^i \vert  z_t^i, \mathcal{M}_{\mathcal{D}_i}^{(k)}, i$)
                    \State $\mathcal{L} \texttt{+=} \mathbb{E}_{E(x), z^{i-1}, E(\hat{x}, \epsilon \sim \mathcal{N}(0,1))}[ \vert  \vert \epsilon - \epsilon'_\psi \vert  \vert _2^2]$ 
                \EndFor
                \State Take a gradient step on: $\nabla_\psi[ \vert  \vert \epsilon - \epsilon'_\psi \vert  \vert _2^2]$
            \EndFor
        \Until{convergence}
    \end{algorithmic}
\end{algorithm}

\begin{algorithm}
    \caption{RISSOLE Sampling}
    \label{algo:sample}
    \begin{algorithmic}[1]
        \State $\hat{z}^0 \sim \mathcal{D}_0$
        \State $\mathcal{M}_{\mathcal{D}}^{(k)} = \xi_k(\hat{z}^0, \mathcal{D})$ where $\mathcal{M}_{\mathcal{D}}^{(k)} = \bigcup_{i=0}^{b-1}\mathcal{M}_{\mathcal{D}_i}^{(k)}$
        \For{each $z^i$}
            \State $z_T^i = \mathcal{N}(0, \mathbf{I})$
            \For{$t = T$ to 1}
                \State $z_{t-1}^i = \frac{1}{\sqrt{1-\beta_t}}(z_t^i - \frac{\beta_t}{\sqrt{1-\alpha_t}}p_\psi(z^i_{t-1} \vert z^i_t, \mathcal{M}_{\mathcal{D}_i}^{(k)}, t, i)) + \sigma_t\epsilon$
            \EndFor
        \EndFor
        \State Reconstruct $z$ as $\bigcup_{i=0}^{b-1} z^i$
        \State $x = D_\phi (z)$
    \end{algorithmic}
\end{algorithm}

\section{Searching $\mathcal{D}$ with entire $z$}
\begin{figure}
    \centering
    \includegraphics[width=\textwidth, keepaspectratio]{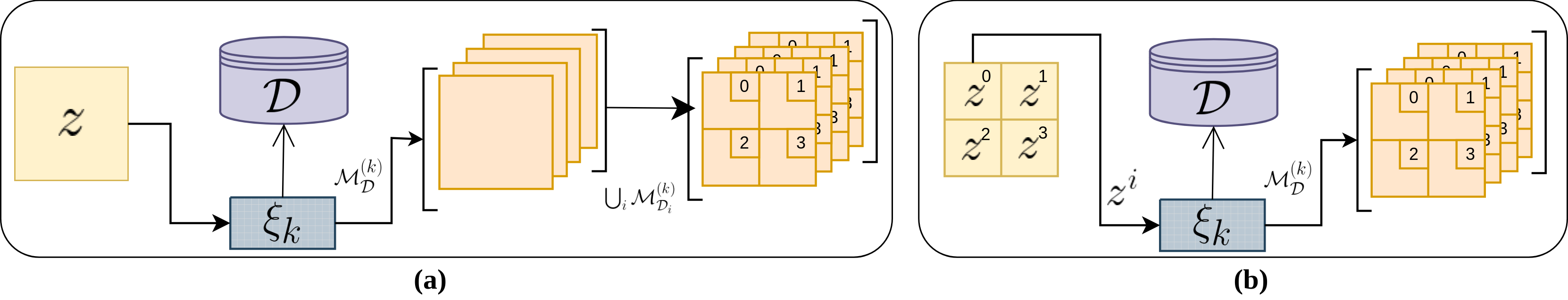}
    \caption{(a) Rather than partitioning $z$ into $\bigcup_{i}z^i$, the entirety of $z$ is fed into $\xi_k(\cdot)$ to obtain $\mathcal{M}_{\mathcal{D}}^{(k)}$. (b) The initial algorithm outlined in the main document involves sending the first segment of $z$, denoted as $z^0$, as a query to locate the nearest neighbors.}
    \label{fig:diag_z_zi}
\end{figure}

As mentioned in Section 4.1.3 in the main paper, the nearest neighbors, $\mathcal{M}_{\mathcal{D}}^{(k)}$ are retrieved from the database $\mathcal{D}$ based on the first block, denoted by $z^0$ of the latent representation $z = E_\theta(x)$ where $x$ is the image and $E_\theta$ is the encoder (in our case, a VQGAN) parameterized by $\theta$.
\begin{figure}[h]
    \centering
    \includegraphics[width=0.5\linewidth, keepaspectratio]{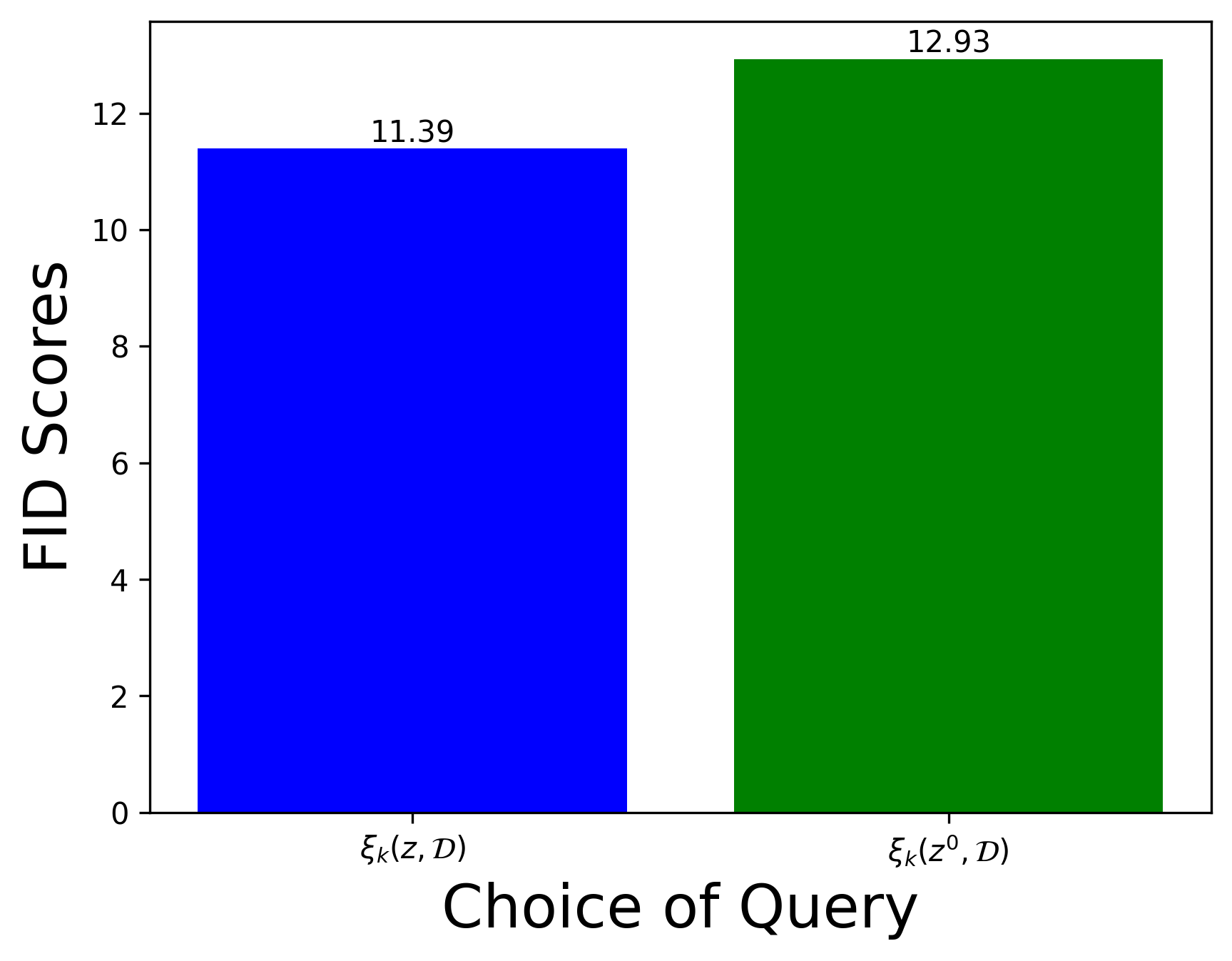}
    \vspace{-1em}
    \caption{{Quantitative evaluation of samples generated from two different RISSOLE models based on $z$ vs. $z^0$.}}
    \label{fig:bar_z_zi}
    \vspace{-1em}
\end{figure}
An alternative design strategy involves leveraging the entirety of $z$ for retrieval, given that we can access the complete image. This slight adjustment in the training process enhances the precision of nearest neighbor searches, resulting in the selection of more pertinent nearest neighbors compared to when the database was queried solely with $z^0$. Figure \ref{fig:diag_z_zi}(a) illustrates the concept of utilizing the entire $z$ within $\xi_k(\cdot)$ for nearest neighbor searches, contrasting with the utilization of only $z^0$ depicted in Figure \ref{fig:diag_z_zi}(b). The disparity in FID scores between RISSOLE models trained and sampled using $z$ and $z^0$ as queries is depicted in Figure \ref{fig:bar_z_zi}. Both models are trained on the CelebA dataset. Evidently, incorporating $\xi_k(z, \mathcal{D})$ results in improved sample generation. Nevertheless, since the search space has expanded by $b$, where $b$ denotes the number of blocks into which the latent space is divided, the memory requirements for constructing the search increase substantially, leading to slightly longer training and sampling times. Depending on the user's preference regarding this trade-off, RISSOLE can be tailored to accommodate either search approach.
\section{Varying the number of blocks in RISSOLE}
The analyses conducted on RISSOLE's samples and FID scores so far have been based on dividing the latent space into four equally sized partitions, denoted as $b = 4$. In this section, we evaluate the qualitative and quantitative sample quality generated by RISSOLE models as the value of $b$ varies among $\{4, 9, 16\}$. This entails examining scenarios where the latent space is divided into grids of dimensions $2 \times 2$, $3 \times 3$, and $4 \times 4$. Since the images
\begin{figure}
    \centering
    \includegraphics[width=0.5\linewidth, keepaspectratio]{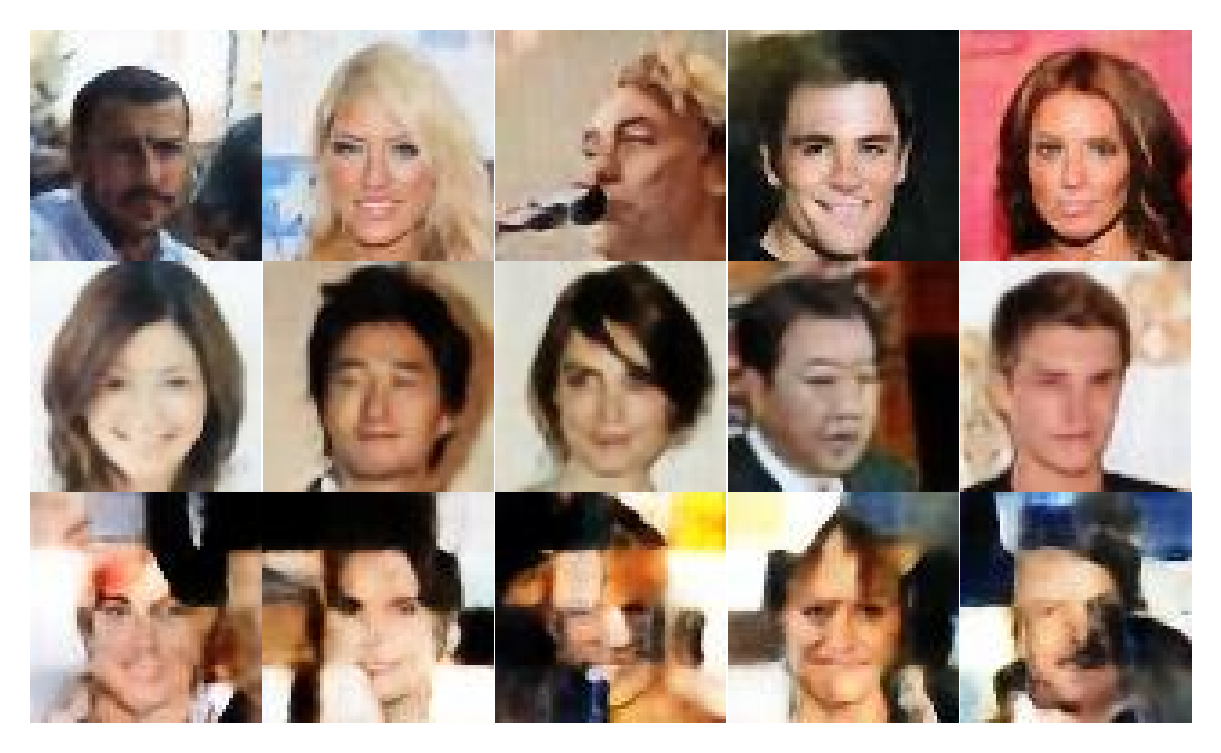}
    \caption{Samples generated by RISSOLE with $b=4$ (top), $b=9$ (middle) and $b=16$ (bottom).}
    \label{fig:b_4_9_16}
\end{figure}
These variations are experimented with using the CelebA dataset. As depicted in Figure \ref{fig:b_4_9_16}, it is evident that RISSOLE performs acceptably with $b=4$ (FID 11.93) and $b=9$ (FID = 12.19). However, noticeable artifacts emerge when $b$ increases to 16 (FID = 97.15). Opting for $b=9$ over 4 reduces GPU usage per training epoch, albeit at the expense of increased training and sampling time, which can be mitigated by training and sampling each block in parallel. These findings are presented based on the RISSOLE variant proposed in the preceding section, which employs a search for nearest neighbors across the entire latent space.